\pdfoutput=1
\PassOptionsToPackage{table}{xcolor}
\documentclass{article}

\usepackage{conference,times}

\usepackage{amsmath,amsfonts,bm}

\def\eqref#1{equation~\ref{#1}}

\def\1{\bm{1}}

\DeclareMathAlphabet{\mathsfit}{\encodingdefault}{\sfdefault}{m}{sl}
\SetMathAlphabet{\mathsfit}{bold}{\encodingdefault}{\sfdefault}{bx}{n}

\definecolor{iccvblue}{rgb}{0.21,0.49,0.74}
\usepackage{tabularx}
\usepackage[breaklinks,colorlinks,allcolors=iccvblue]{hyperref}
\usepackage{graphicx}
\usepackage{colortbl}
\definecolor{mygray}{gray}{0.92}
\definecolor{mygreen}{RGB}{220, 245, 220}

\usepackage{array}

\definecolor{medicalgray}{RGB}{242,242,242}
\definecolor{oursgreen}{RGB}{218,245,226}
 
\usepackage{url}
\usepackage{amsmath}
\usepackage{amssymb}
\usepackage{booktabs}
\usepackage{multirow}
\usepackage{tcolorbox}
\definecolor{best}{HTML}{C8E6C9}
\definecolor{best2}{HTML}{BBDEFB}
\definecolor{grey}{HTML}{e0e0e0}

\usepackage{algorithm}
\usepackage{algorithmic}
\usepackage{pifont}
\usepackage{cuted} 
\usepackage{newfloat}
\usepackage{listings}
\usepackage{soul}
\usepackage{marvosym}
\usepackage{wrapfig}
\usepackage{caption}  

\definecolor{oursgreen}{RGB}{218,245,226}
\newcommand{\tablestyle}{\footnotesize\setlength{\tabcolsep}{4pt}\renewcommand{\arraystretch}{1.1}}

\usepackage{tocloft}
\usepackage{etoc}
\usepackage{enumitem}
\usepackage[T1]{fontenc} 
\usepackage{float}

\title{X-Ragent: Tool-Augmented Reinforcement Learning for Evidence-Grounded Chest X-Ray Reasoning}

\author{
\textbf{
Yi Wang\textsuperscript{1 2 *} \quad
Zhenlong Yuan\textsuperscript{1 2 * \ensuremath{\ddagger}} \quad
Yuchan Guo\textsuperscript{3 *} \quad
Maolei Huang\textsuperscript{4}
}
\\[0.6em]
\textbf{
Zijian Song\textsuperscript{5} \quad
Dapeng Zhang\textsuperscript{6} \quad
Rongbin Tan\textsuperscript{1} \quad
Jiyuan Wang\textsuperscript{7}
}
\\[0.6em]
\textbf{
Zhiyuan Wang\textsuperscript{8} \quad
Fangfang Lin\textsuperscript{9} \quad
Shuhan Song\textsuperscript{1 2 \textsection} \quad
Huawei Cao\textsuperscript{1 2}
}
\\[1.0em]
\textsuperscript{1} Institute of Computing Technology \quad
\textsuperscript{2} University of Chinese Academy of Sciences 
\\[0.6em]
\textsuperscript{3} Carnegie Mellon University \quad
\textsuperscript{4} Alibaba Group \quad
\textsuperscript{5} Sun Yat-sen University \quad
\\[0.6em]
\textsuperscript{6} National University of Singapore \quad
\textsuperscript{7} Beijing Jiaotong University
\\[0.6em]
\textsuperscript{8} City University of Hong Kong \quad
\textsuperscript{9} Santa Clara University
\\[1.0em]
\textsuperscript{*}~\textit{Equal contribution}\quad
\textsuperscript{\ensuremath{\ddagger}}~\textit{Project Lead}\quad
\textsuperscript{\textsection}~\textit{Corresponding Author}
}

\iclrfinalcopy 
\begin{document}

\maketitle

\begin{abstract}
Chest X-ray (CXR) interpretation often requires integrating visual findings with task-specific evidence rather than relying on direct visual recognition alone. While tool-augmented medical agents can supply such evidence via external modules, they often lack explicit mechanisms for learning when auxiliary evidence is needed and for preventing inapplicable quantitative evidence from entering diagnostic reasoning. To address this, we propose \textbf{X-Ragent}, a post-training framework for tool-augmented reinforcement learning that decouples learned auxiliary-evidence seeking from validity-gated quantitative measurement acquisition for evidence-grounded CXR reasoning. We construct \textbf{Rad-CoT}, a 7.7K-sample corpus combining causally ordered tool-augmented trajectories with report-style diagnostic reasoning, and use it to initialize Qwen3-VL-8B via agentic supervised fine-tuning (SFT). We then optimize the policy with Group Relative Policy Optimization (GRPO) using an accuracy-gated reward that applies process-level terms for measurement consistency and auxiliary-tool cost only to correct trajectories, preventing plausible process behavior from compensating for an incorrect diagnosis. Across three CXR benchmarks, X-Ragent achieves the strongest results, including an 8.6-point improvement on the fully held-out ChestAgentBench. By seeking evidence selectively and admitting only valid measurements, X-Ragent supports more faithful and evidence-grounded CXR reasoning.

\end{abstract}

\section{Introduction}
\begin{figure}[t]
    \centering
    \includegraphics[width=\linewidth]{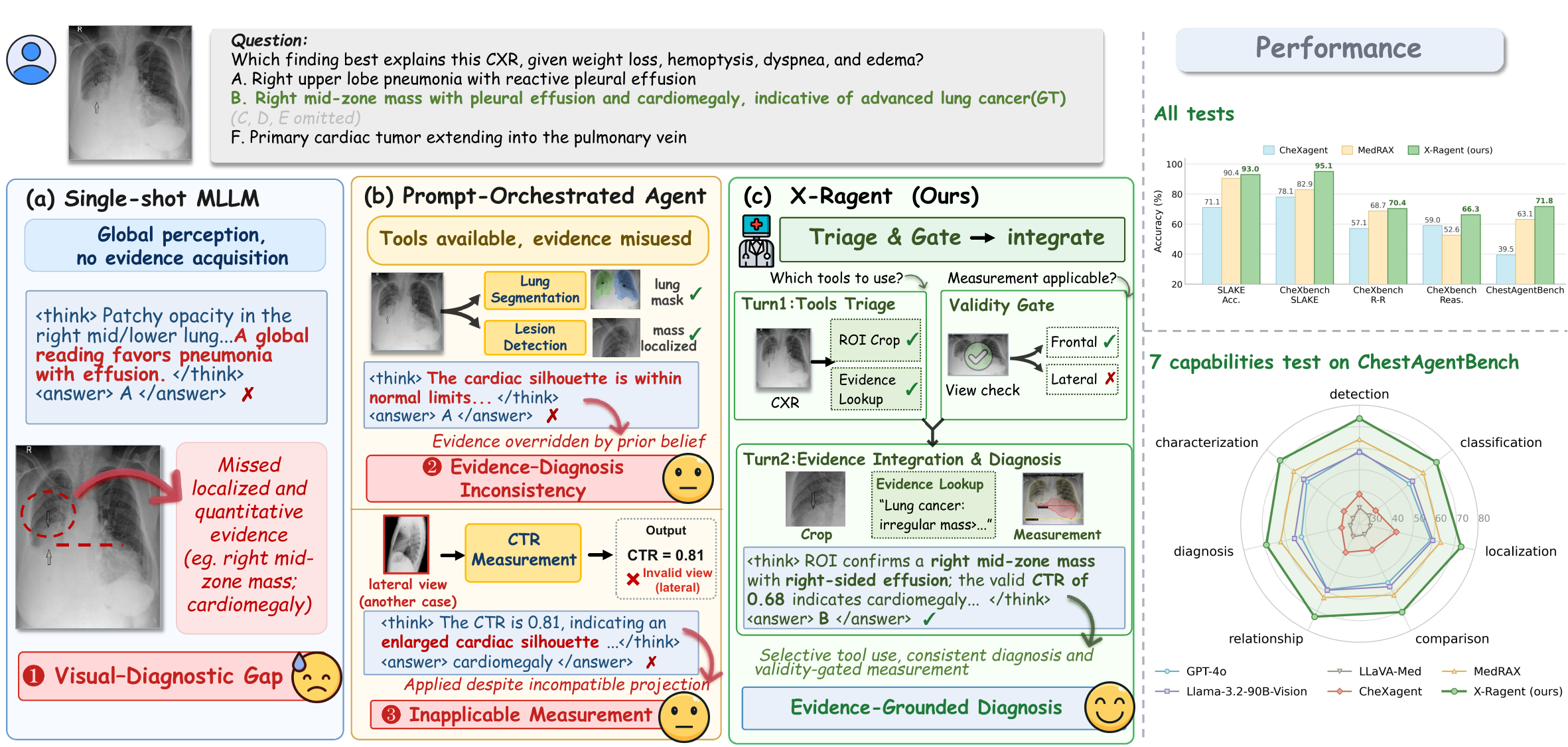}
    \caption{\textbf{Evidence-grounded CXR reasoning with X-Ragent.}
(a) Single-shot MLLMs read the radiograph globally and miss localized evidence, leaving a visual--diagnostic gap.
(b) Tool-augmented agents invoke tools but override the returned evidence with textual priors (top), or apply a measurement under incompatible acquisition conditions (bottom).
(c) X-Ragent routes task-specific evidence on demand and integrates it with validity-gated quantitative measurements for evidence-grounded diagnosis.}
    \label{fig:intro}
    \vspace{-0.25in}  
\end{figure}
Chest X-ray (CXR) is the most frequently ordered imaging examination, supporting clinical tasks from thoracic disease detection to postoperative monitoring~\citep{yang2025chexfound}.
Automated CXR analysis has progressed from closed-set disease classification~\citep{ma2025arkplus} to report generation~\citep{jin2024promptmrg, liu2025mlrg} and visual question answering~\citep{hu2024medicalcxrvqa}, and is now moving toward open-ended interpretation with explanations~\citep{chen2024chexagent, liu2025gemex}.
However, reliable CXR reasoning hinges on localized and quantitative evidence, which current vision-language models still handle poorly. They struggle to ground diagnostic criteria in the relevant anatomy~\citep{lee2025cxreasonbench} and frequently hallucinate measurements~\citep{heiman2025factchexcker}.

Recent advances make such evidence-grounded reasoning increasingly feasible.
General and medical multimodal large language models (MLLMs)~\citep{bai2025qwen3, park2025m4cxr} provide strong vision-language backbones, and GRPO-based reinforcement learning effectively elicits multi-step reasoning, including in CXR~\citep{shao2024deepseekmath, chestxreasoner2025}.
Meanwhile, specialized models supply fine-grained visual evidence through anatomical segmentation~\citep{seibold2023cxas} and text-prompted lesion localization~\citep{zhao2024biomedparse}, which tool-augmented agents compose into diagnostic workflows~\citep{fallahpour2025medrax, li2024mmedagent}.

Despite these advances, three gaps remain when current approaches are applied to complex CXR diagnostic tasks:
\ding{182} \textbf{Visual-Diagnostic Gap}: MLLMs globally perceive radiographs but struggle to link broad visual cues with localized, fine-grained anatomical evidence or quantitative metrics. This stems from a lack of structured cognitive mapping connecting perception to diagnostic hypotheses, leading to fragmented reasoning.
\ding{183} \textbf{Evidence–Diagnosis Inconsistency}: Existing medical agents may invoke external tools but frequently ignore or misinterpret their objective outputs. Driven by textual priors rather than grounded evidence, they generate plausible-sounding but unsupported diagnoses, suffering from imprecise alignment between visual findings and semantic conclusions.
\ding{184} \textbf{Inapplicable Measurement}: Deterministic measurements are meaningful only under their intended acquisition conditions. Without an explicit applicability check, inapplicable quantitative evidence can enter the reasoning process and mislead the final diagnosis.

To resolve these limitations, we argue that an effective CXR diagnostic agent requires three synergistic capabilities:
\ding{182} \textbf{Evidence-Triage Planning}: The capability to assess whether auxiliary evidence is needed and selectively route task-specific evidence accordingly. Conditioned on the clinical query and initial radiographic assessment, the policy invokes visual refinement or condition-specific radiographic knowledge only when additional evidence is needed, alleviating the visual-diagnostic gap.
\ding{183} \textbf{Faithful-Evidence Integration}: The capability to reason over heterogeneous tool observations and reconcile them with visual findings. This evidence-grounded process helps the agent resolve visual-semantic ambiguities rather than blindly trusting tool outputs, thereby mitigating cross-modal hallucination.
\ding{184} \textbf{Applicability-Gated Measurement}: A validity gate that automatically provides basic CXR measurements only when acquisition conditions are satisfied, preventing incompatible quantitative evidence from entering diagnostic reasoning.

To realize these capabilities, we introduce \textbf{X-Ragent}, a post-training framework for evidence-grounded CXR reasoning. As shown in Fig.~\ref{fig:intro}, X-Ragent selectively acquires task-relevant evidence, gates quantitative measurements by clinical applicability, and integrates the resulting evidence into diagnostic reasoning. First, we construct the \textbf{Rad-CoT} dataset for \textbf{supervised fine-tuning (SFT)}, comprising structured diagnostic trajectories that teach the policy when to invoke optional visual-refinement and knowledge-lookup tools and how to integrate their observations into diagnostic reasoning. Second, quantitative CXR evidence is supplied through a deterministic, validity-gated pathway: measurements such as the cardiothoracic ratio are exposed only when the applicability check succeeds. Finally, we optimize the policy with Group Relative Policy Optimization (GRPO) using a composite \textbf{gated reward} that prioritizes diagnostic correctness while encouraging selective auxiliary-tool use and faithful interpretation of valid quantitative evidence. This design decouples learned evidence seeking from validity-constrained measurement acquisition: X-Ragent decides \emph{which} optional evidence to seek, while the validity gate determines \emph{whether} quantitative measurements are admitted into reasoning. Our contributions are summarized as follows:

\begin{itemize}[leftmargin=*,nosep]
    \item \textbf{Evidence-Grounded CXR Reasoning.}
    We propose \textbf{X-Ragent}, a post-training framework that decouples learned auxiliary-evidence seeking from validity-gated quantitative measurement, enabling selective and evidence-grounded CXR reasoning.

    \item \textbf{Structured Agentic Post-Training.}
    We construct \textbf{Rad-CoT}, a 7.7K-sample dataset of structured diagnostic trajectories, and develop a two-stage SFT--GRPO training scheme with an accuracy-gated reward that promotes selective tool use and faithful interpretation of verifiable radiographic measurements.

    \item \textbf{Empirical Effectiveness.}
    X-Ragent achieves state-of-the-art performance across three CXR benchmarks, raising accuracy on the fully held-out \textbf{ChestAgentBench} from 63.1\% to 71.7\% over the strongest baseline, with consistent gains across all seven CXR interpretation capabilities.
\end{itemize}

\section{Methodology}
\begin{figure}[t]
    \centering
    \includegraphics[width=\linewidth]{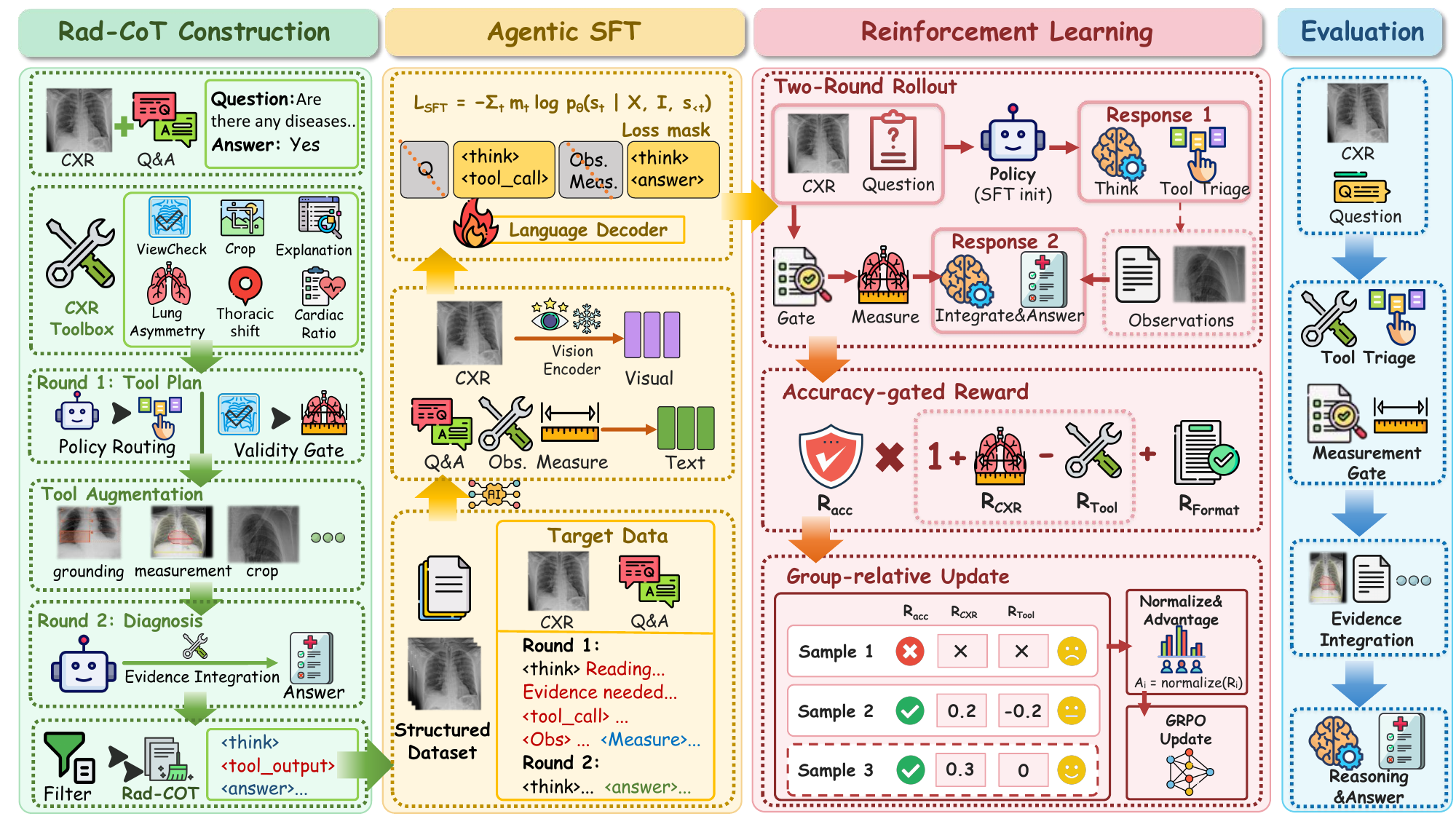}
    \caption{\textbf{Overview of X-Ragent.}
(i) Rad-CoT constructs structured CXR reasoning data through evidence triage and gating, evidence integration, and filtering.
(ii) Agentic SFT cold-starts Qwen3-VL-8B on Rad-CoT, masking injected observations from the language-modeling loss.
(iii) GRPO optimizes the policy with accuracy and format supervision, while tool-use cost and radiographic evidence consistency are gated by answer correctness.
(iv) At inference, the policy selectively routes auxiliary tools, while an independent validity gate decides whether quantitative measurements are exposed; both are integrated in the second turn.}
    \label{fig:framework}
    \vspace{-0.25in}
\end{figure}
\textbf{Overview.}
As illustrated in Fig.~\ref{fig:framework}, we first formulate tool-augmented CXR reasoning (Sec.~\ref{sec:Problem}), then construct Rad-CoT (Sec.~\ref{sec:data}), and post-train the policy via agentic SFT (Sec.~\ref{sec:sft}) and GRPO with a gated reward (Sec.~\ref{sec:rl}).

\subsection{Problem Formulation \& Workflow Design}
\label{sec:Problem}
\textbf{Problem Formulation.} We formulate chest X-ray (CXR) interpretation as a tool-augmented visual reasoning process. Given a CXR image $I \in \mathbb{R}^{H \times W \times 3}$ and a clinical query $Q$, the model is required to generate a structured output $O = \langle \mathcal{R}, A \rangle$, where $\mathcal{R}$ denotes the reasoning trajectory including tool invocations and their observations, and $A$ denotes the final answer. The answer may take the form of a multiple-choice selection, a short-answer response, or an image--text reasoning judgment, depending on the downstream task. Instead of directly mapping the image to a prediction, we instantiate Qwen3-VL-8B as an agentic policy $\pi_\theta$. The policy selectively invokes auxiliary tools from an action set $\mathcal{T}_{\pi}$, while deterministic quantitative measurements are supplied through an independent validity-gated path.

\textbf{Agentic Inference.}
X-Ragent performs diagnosis through a two-turn autoregressive process that separates policy-routed auxiliary acquisition from validity-gated quantitative measurement.

\textbf{Stage 1: Evidence Triage.}
Given an image $I$ and query $Q$, the policy first performs a global radiographic assessment $z_1$ and selects an optional auxiliary-tool plan $C$, a possibly empty sequence of calls from $\mathcal{T}_{\pi}=\{\texttt{region\_parse},\texttt{roi\_crop},\texttt{evidence\_lookup}\}$ with their arguments (e.g., target region or candidate conditions):
\begin{equation}
(z_1, C)\sim\pi_\theta(\cdot\mid I,Q;\mathcal{T}_{\pi}),\qquad C=(c_1,\dots,c_K),\ c_k\in\mathcal{T}_{\pi},\ 0\le K\le K_{\max},
\end{equation}
where $K=0$ when visual inspection alone suffices and $K_{\max}=5$. Independently of this routing decision, a validity gate $g(I)\in\{0,1\}$ exposes the quantitative measurement bundle only when the validity gate passes.

\textbf{Stage 2: Evidence Integration and Diagnosis.}
The routed calls in $C$ are executed and their observations are returned to the policy; when $g(I)=1$, the backend additionally supplies the quantitative measurement bundle. Conditioned on its first-turn output, the policy integrates these observations with the original image and query to generate the reasoning trajectory $\mathcal{R}$ and final answer $A$:
\begin{equation}
O = \langle \mathcal{R}, A \rangle
\sim
\pi_\theta
\left(
\cdot \mid I \oplus F_C,\,
Q \oplus z_1 \oplus C \oplus E_C \oplus M(I)
\right),
\end{equation}
where $\oplus$ denotes appending to the multimodal context, $F_C$ and $E_C$ denote visual and semantic feedback from the routed calls, respectively, and $M(I)$ denotes the quantitative measurement bundle, with $M(I)=\varnothing$ when $g(I)=0$.

\begin{figure}[t]
    \centering
    \includegraphics[width=\linewidth]{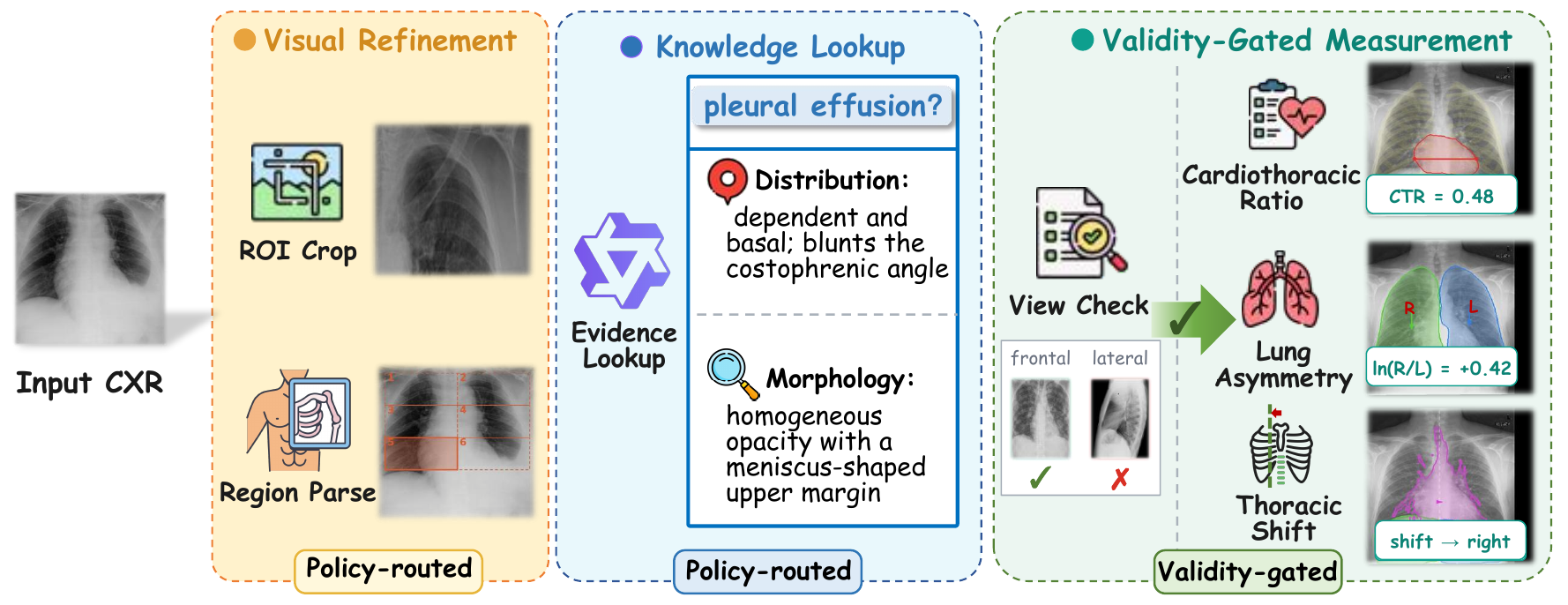}
    \caption{\textbf{X-Ragent tool suite.}
    The policy selectively invokes visual-refinement and evidence-lookup tools,
    while a separate validity-gated backend provides quantitative measurements.}
    \label{fig:tool}
\end{figure}

\textbf{Tool Suite.}
As shown in Fig.~\ref{fig:tool}, the policy can invoke three auxiliary tools:
\texttt{roi\_crop} magnifies relevant regions,
\texttt{region\_parse} supports localized anatomical inspection, and
\texttt{evidence\_lookup} retrieves condition-specific radiographic knowledge from a locally served Qwen3.5-27B. Separately, a deterministic backend computes the cardiothoracic ratio, lung-field asymmetry, and thoracic shift, exposing them only when \texttt{view\_check} confirms applicability.
Only these verifiable measurements receive explicit process supervision; further details are provided in Appendix~\ref{appx:tool_library}.

\subsection{Training Data Construction}
\label{sec:data}
Existing MLLM training data mainly supervise final answers, without capturing how diagnostic evidence is acquired and integrated into reasoning. We therefore construct \textbf{Rad-CoT}, a 7,715-sample SFT corpus comprising 3,715 routed two-turn examples and 4,000 single-turn report-style CoT examples.

\textbf{Data Collection.}
Rad-CoT combines PA radiographs from MIMIC-CXR-JPG~\citep{PhysioNet-mimic-cxr-jpg-2.1.0,pollard_physionet_2026} with structured findings derived from LLaVA-Rad~\citep{PhysioNet-llava-rad-mimic-cxr-annotation-1.0.0,pollard_physionet_2026} annotations and publicly available medical VQA data. It contains 1,995 multiple-choice examples, 720 closed-ended VQA examples, 1,000 binary-choice VQA examples, and 4,000 report-style diagnostic CoT examples.

\textbf{Generating Pipeline.}
We use locally deployed Qwen3.5-122B-A10B-FP8 to construct Rad-CoT reasoning data, with routed examples following the two-turn protocol in Sec.~\ref{sec:Problem}:

\begin{itemize}[leftmargin=*,nosep]
\item \textbf{Evidence Triage and Acquisition.}
For routed examples, the first turn performs an initial radiographic assessment and selects optional visual-refinement or knowledge-lookup actions, while quantitative measurements are exposed only when \texttt{view\_check} verifies their applicability.

\item \textbf{Evidence Integration.}
The second turn reconciles the returned observations with the original radiograph to produce the final reasoning and answer. Observations are revealed only after the corresponding action is generated, preserving inference-time causal ordering.

\item \textbf{Trajectory Filtering.}
We filter malformed, leaky, or unverifiable generations; for the report-style subset, this reduces approximately 6{,}000 candidates to 4{,}000 valid examples. We also remove report-derived label leakage from multiple-choice trajectories.
\end{itemize}

Construction and quality-control details are provided in Appendix~\ref{app:data_construction}.

\subsection{Agentic Supervised Fine-Tuning}
\label{sec:sft}
After constructing Rad-CoT, we adopt SFT as a cold start before GRPO.
Prompt-elicited reasoning underperforms direct answering
(Appendix~\ref{app:prompt_reasoning}), while direct GRPO from the base policy
also underperforms supervised initialization (Tab.~\ref{tab:pipeline}).
We therefore first use Rad-CoT to establish structured reasoning and tool-use
behaviors before RL optimization.

Formally, each training instance is denoted as $\mathcal{T} = (\mathcal{X}, \mathcal{I}, \mathcal{S}, \mathcal{Y})$, where $\mathcal{X}$ denotes the visual input, $\mathcal{I}$ is the clinical task instruction, $\mathcal{S} = \{s_1, \dots, s_T\}$ represents the policy-generated tokens spanning both the routing decision and the reasoning trajectory, and $\mathcal{Y}$ denotes the final answer. We apply a selective masking strategy: the loss is computed exclusively over policy-generated tokens, with injected observation spans masked out. The objective minimizes:
\begin{equation}
\label{eq:sft}
\mathcal{L}_{\text{SFT}} = -\mathbb{E}_{\mathcal{T} \sim \mathcal{D}} \left[ \sum_{t=1}^{T} \log p_\theta(s_t \mid \mathcal{X}, \mathcal{I}, s_{<t}) \right],
\end{equation}
where $p_\theta(\cdot)$ denotes the conditional probability distribution of the parameterized policy. This phase anchors auxiliary-tool routing and evidence-integrated diagnostic reasoning, providing a calibrated policy initialization for subsequent reinforcement learning.

\subsection{Agentic Reinforcement Learning}
\label{sec:rl}
\textbf{Group Relative Policy Optimization (GRPO).}
We adopt the GRPO algorithm~\citep{shao2024deepseekmath} for policy optimization. 
GRPO innovatively employs a groupwise comparison framework to evaluate candidate responses.
Specifically, for each query $ q $ paired with its ground-truth solution $ a $ from dataset $ D $, the algorithm generates a set of rollout trajectories $ \{o_1, o_2, \dots, o_G\} $ based on the previous policy $ \pi_{\theta_{\text{old}}} $. The policy $ \pi_\theta $ is then refined through the optimization of this objective function:  
\begin{equation}
\begin{aligned}
    &\mathcal{L}_{GRPO}(\theta) = -\mathbb{E}_{q \sim P(Q), \{o_i\}_{i=1}^{G} \sim \pi_{\theta_{old}}(O|q)} \Bigg[ \frac{1}{G} \sum_{i=1}^G \\
    &\quad \Bigg( \min\left( \frac{\pi_\theta(o_i|q)}{\pi_{\theta_{old}}(o_i|q)} A_i, \text{clip}\left(\frac{\pi_\theta(o_i|q)}{\pi_{\theta_{old}}(o_i|q)}, 1-\epsilon, 1+\epsilon\right) A_i \right) - \beta \mathbb{D}_{KL}(\pi_\theta \| \pi_{ref}) \Bigg) \Bigg],
\end{aligned}
\end{equation}
\begin{equation}
\mathbb{D}_{KL}(\pi_\theta||\pi_{ref}) = \frac {\pi_{ref}(o_i|q)} {\pi_\theta(o_i|q)} - \log \frac {\pi_{ref}(o_i|q)} {\pi_\theta(o_i|q)} - 1,
\end{equation}
where $ \beta $ is adopted to balance the trade-off between exploration and stability during optimization. Then the advantage estimator $ A_{i} $ is calculated using normalized rewards from the trajectory group:  
\begin{equation}
A_{i} = \frac{r_i - \text{mean}(\{r_1, r_2, \dots, r_G\})}{\text{std}(\{r_1, r_2, \dots, r_G\})}.
\end{equation}
Each trajectory $ o_i $ receives a composite reward through rule-based verification designed as follows.

\textbf{Reward Design.}
Outcome-level rewards supervise final diagnostic correctness but provide limited guidance on whether acquired evidence is interpreted consistently. We therefore retain answer accuracy as the primary optimization target while introducing lightweight process-level signals for evidence consistency and auxiliary-tool efficiency. The reward comprises answer accuracy $R_{\mathrm{acc}}$, format compliance $R_{\mathrm{format}}$, radiographic evidence consistency $R_{\mathrm{CXR}}$, and tool-use cost $R_{\mathrm{tool}}$. $R_{\mathrm{acc}}\in\{0,1\}$ evaluates final-answer correctness, while $R_{\mathrm{format}}\in\{0,-1\}$ penalizes malformed outputs.

$R_{\mathrm{CXR}}$ applies only to objective measurements whose applicability and interpretation can be deterministically verified. It rewards consistency with the query-relevant measurements exposed by the validity-gated backend, where query relevance affects reward eligibility only, not measurement execution. In parallel, $R_{\mathrm{tool}}$ counts policy-generated auxiliary calls in $\mathcal{T}_{\pi}$, excluding backend validity checks and deterministic measurement routines. Detailed definitions are provided in Appendix~\ref{app:reward}.
The total reward is
\begin{equation}
\label{eq:total_reward}
R(\tau)=R_{\mathrm{format}}(\tau)+R_{\mathrm{acc}}(\tau)\left[1+\lambda_mR_{\mathrm{CXR}}(\tau)-\lambda_tR_{\mathrm{tool}}(\tau)\right].
\end{equation}
The multiplicative accuracy gate restricts the process-level terms to refining preferences among correct trajectories. Without this gate, an incorrect trajectory could still gain reward from superficially desirable process behavior, allowing process quality to compensate for diagnostic error. Visual-refinement and knowledge-lookup tools receive no explicit process supervision and are optimized indirectly through task success.

\section{Experiment}
\textbf{Benchmarks and protocol.}
We evaluate X-Ragent on three CXR benchmarks, with further details provided in Appendix~\ref{appx:exp_details}.
ChestAgentBench~\citep{fallahpour2025medrax} contains 2,500 six-choice questions spanning seven capabilities and is entirely excluded from both SFT and RL training, serving as an unseen benchmark.
For CheXbench~\citep{chen2024chexagent}, following MedRAX~\citep{fallahpour2025medrax}, we use 238 VQA questions from Rad-ReStruct~\citep{pellegrini2023rad} and SLAKE and 380 fine-grained reasoning questions from OpenI~\citep{demner2016preparing}.
SLAKE~\citep{liu2021slake} provides 114 closed-ended CXR test questions, scored by exact match following M4CXR~\citep{park2025m4cxr}.
For CheXbench and SLAKE, training uses only the official training partitions of their source datasets, and all evaluation images are excluded from training (Appendix~\ref{app:data_construction}).

\textbf{Implementation Details.}
X-Ragent is initialized from Qwen3-VL-8B-Instruct and post-trained with SFT followed by GRPO.
For SFT, we train on 7.7K samples for one epoch with a learning rate of $1\times10^{-5}$ and an effective batch size of 16.
For GRPO, we optimize the SFT checkpoint on 4.7K training examples for 300 steps, using 4 rollouts per example and at most 5 policy-routed auxiliary calls per trajectory.
We use a learning rate of $2\times10^{-6}$, a rollout batch size of 16, and an actor update batch size of 8.
Additional training details are provided in Appendix~\ref{appx:exp_details}.

\subsection{Main Results}
\label{sec:main_results}

\begin{table}[t]
\centering
\caption{\textbf{Main results on three CXR benchmarks.} Accuracy (\%). CheXbench reports Rad-Restruct (R-R), SLAKE and OpenI reasoning (Reas.); ChestAgentBench is fully held out (OOD) and reported per capability. Following MedRAX, ChestAgentBench ``All'' is question-level accuracy over all 2,500 questions; since capability labels overlap, it is not the mean of the category columns. $^\dagger$Our base model, evaluated zero-shot with the direct-answer prompt (Appendix~\ref{app:prompt_reasoning}). We denote \textbf{best} and \underline{second-best} results, and highlight \colorbox{oursgreen}{ours}.}
\label{tab:main_results}
\tablestyle
\setlength{\tabcolsep}{3.2pt}
\begin{tabular}{@{}l ccc c cccccccc@{}}
\toprule
\multirow{2}{*}{\textbf{Model}} & \multicolumn{3}{c}{\textbf{CheXbench}} & \textbf{SLAKE} & \multicolumn{8}{c}{\textbf{ChestAgentBench} (OOD)} \\
\cmidrule(lr){2-4} \cmidrule(lr){5-5} \cmidrule(l){6-13}
& R-R & SLAKE & Reas. & Acc. & Det. & Cls. & Loc. & Comp. & Rel. & Diag. & Char. & All \\
\midrule
\multicolumn{13}{@{}l}{\textit{General-purpose VLMs}} \\
GPT-4o               & 53.9 & \underline{85.4} & 51.1 & -- & 58.7 & 54.6 & 59.0 & 55.5 & 59.0 & 52.6 & 56.1 & 56.4 \\
Llama-3.2-90B-Vision & 62.6 & 74.0 & 49.2 & -- & 58.1 & 56.5 & 59.9 & 57.5 & 59.3 & 55.9 & 58.0 & 57.9 \\
Qwen3-VL-8B$^\dagger$ & \textbf{70.4} & 75.6 & 50.5 & 72.8 & 56.5 & 53.3 & 57.9 & 55.9 & 57.2 & 52.1 & 53.8 & 55.1 \\
\midrule
\multicolumn{13}{@{}l}{\textit{Medical VLMs}} \\
LLaVA-Med            & 34.9 & 55.5 & 45.8 & -- & 32.4 & 30.8 & 30.2 & 30.6 & 31.8 & 29.3 & 28.8 & 28.7 \\
CheXagent            & 57.1 & 78.1 & \underline{59.0} & 71.1 & 38.7 & 34.7 & 42.5 & 38.5 & 39.8 & 33.5 & 34.2 & 39.5 \\
\midrule
\multicolumn{13}{@{}l}{\textit{Agentic CXR systems}} \\
MedRAX & \underline{68.7} & 82.9 & 52.6 & \underline{90.4} & \underline{64.1} & \underline{62.9} & \underline{63.6} & \underline{61.8} & \underline{63.1} & \underline{62.5} & \underline{64.0} & \underline{63.1} \\
\rowcolor{oursgreen}
\textbf{X-Ragent (Ours)} & \textbf{70.4} & \textbf{95.1} & \textbf{66.3} & \textbf{93.0} & \textbf{73.8} & \textbf{70.7} & \textbf{73.3} & \textbf{70.5} & \textbf{72.9} & \textbf{69.3} & \textbf{72.1} & \textbf{71.7} \\
\bottomrule
\end{tabular}
\end{table}

As shown in Table~\ref{tab:main_results}, X-Ragent achieves state-of-the-art performance across all three benchmarks, reaching 71.7\% on ChestAgentBench, averaging 77.3\% across the three CheXbench subsets, and 93.0\% on SLAKE. We highlight three observations.
\ding{182}~\textbf{Strong performance across diverse CXR tasks.} X-Ragent improves over the best reported results by 8.6, 9.2, and 2.6 points on ChestAgentBench, CheXbench, and SLAKE, respectively. Relative to its Qwen3-VL-8B base model, post-training yields gains of 16.6, 11.8, and 20.2 points, enabling an 8B model to surpass substantially larger general-purpose VLMs and the GPT-4o-based MedRAX agent. On CheXbench, the gains come from the SLAKE (+19.5) and OpenI reasoning (+15.8) subsets, whereas on the 115-question Rad-Restruct subset X-Ragent matches its base model (70.4\%).
\ding{183}~\textbf{Strong out-of-domain generalization.} ChestAgentBench is entirely excluded from post-training, yet X-Ragent outperforms the strongest baseline across all seven interpretation capabilities in Table~\ref{tab:main_results} and improves over its base model by 14.6--18.3 points on every capability. This suggests that the learned reasoning behaviors transfer beyond the benchmarks represented during optimization.
\ding{184}~\textbf{Learned and selective auxiliary-evidence routing.} X-Ragent learns when to invoke optional visual-refinement and knowledge-lookup tools under GRPO, while quantitative measurements follow a separate validity-gated path. Tab.~\ref{tab:need_stratified} shows that acquisition is conditioned on evidence need rather than applied by default.
Sec.~\ref{sec:ablation} further analyzes the contributions of tool-augmented training, evidence-consistency supervision, and applicability gating.

\subsection{Ablation Study}
\label{sec:ablation}

\begin{table}[t]
\centering
\begin{minipage}[t]{0.49\linewidth}
\centering
\caption{\textbf{Post-training pipeline ablation.} ChestA.\ is OOD; CheXb.\ and SLAKE are ID test splits. Avg.\ is the mean over the three benchmarks.}
\label{tab:pipeline}
\tablestyle
\setlength{\tabcolsep}{2pt}
\begin{tabular}{@{}lcccc@{}}
\toprule
Method & ChestA. & CheXb. & SLAKE & Avg. \\
\midrule
(a) Zero-shot       & 55.1 & 65.5 & 72.8 & 64.5 \\
(b) RL only         & 59.8 & 68.4 & 78.9 & 69.0 \\
(c) SFT only        & 64.4 & 72.0 & 89.5 & 75.3 \\
(d) SFT+RL          & 67.4 & 73.2 & 92.1 & 77.6 \\
\rowcolor{oursgreen}
(e) SFT+RL+Tools    & \textbf{71.7} & \textbf{77.3} & \textbf{93.0} & \textbf{80.7} \\
\bottomrule
\end{tabular}
\end{minipage}%
\hfill
\begin{minipage}[t]{0.49\linewidth}
\centering
\caption{\textbf{Reward design ablation (Eq.~\ref{eq:total_reward}).} $R_{acc}$ is always used; rows (c)--(e) add process terms additively, while (f) gates them by $R_{acc}$.}
\label{tab:reward}
\tablestyle
\setlength{\tabcolsep}{2pt}
\begin{tabular}{@{}lccccccc@{}}
\toprule
& $R_{\text{fmt}}$ & $R_{\text{tool}}$ & $R_{\text{CXR}}$ & Gate & ChestA. & CheXb. & SLAKE \\
\midrule
(a) &            &            &            &            & 67.2 & 73.8 & 90.4 \\
(b) & \checkmark &            &            &            & 69.4 & 75.3 & 91.2 \\
(c) & \checkmark & \checkmark &            &            & 70.1 & 75.9 & 92.1 \\
(d) & \checkmark &            & \checkmark &            & 70.6 & 76.4 & 92.1 \\
(e) & \checkmark & \checkmark & \checkmark &            & 69.8 & 75.6 & 91.2 \\
\rowcolor{oursgreen}
(f) & \checkmark & \checkmark & \checkmark & \checkmark & \textbf{71.7} & \textbf{77.3} & \textbf{93.0} \\
\bottomrule
\end{tabular}
\end{minipage}
\end{table}

\textbf{Training pipeline.} In rows (a)--(e) of Tab.~\ref{tab:pipeline}, we compare
zero-shot, RL-only, SFT-only, and the two-stage model evaluated with and without
tool access. SFT in row (c) yields the largest single gain (+10.8), improving both in-domain SLAKE (+16.7) and held-out ChestAgentBench (+9.3), whereas RL in row (d) adds more on held-out ChestAgentBench (+3.0) than on in-domain CheXbench (+1.2). This suggests that SFT establishes the core reasoning behavior, whereas RL mainly improves its transfer. RL-only training in row (b) falls 6.3 points below SFT-only,
which we attribute to the difficulty of eliciting structured reasoning from reward alone, motivating our cold-start design. Enabling tools in row (e) adds 4.3 points on ChestAgentBench and 4.1 on CheXbench, but only 0.9 on SLAKE, where accuracy is already near saturation at 92.1\% and the 114-question test set makes a single item worth 0.88\%. The gains are therefore concentrated on the two benchmarks that involve multi-step and fine-grained reasoning, where external evidence is most likely to be informative.

\textbf{Reward design.} Tab.~\ref{tab:reward} isolates each reward component. $R_{acc}$ alone performs worst on all three benchmarks. The tool-use cost (c) and the evidence reward (d) each improve over (b), but combining them without the gate (e) underperforms either alone: ungated, incorrect trajectories can still earn $R_{\text{CXR}}$, while the tool cost penalizes evidence acquisition regardless of outcome. Gating both terms by answer correctness (f) gives the best results on all three benchmarks, supporting the multiplicative form of Eq.~\ref{eq:total_reward}.

\begin{table}[t]
\centering
\setlength{\abovecaptionskip}{0pt}
\setlength{\belowcaptionskip}{0pt}
\begin{minipage}[t]{0.47\textwidth}
\caption{\textbf{Selective auxiliary-tool routing.}
Questions are stratified by base-model correctness; TIR (tool invocation rate) and Acc.\ are those of X-Ragent.}
\label{tab:need_stratified}
\end{minipage}%
\hfill
\begin{minipage}[t]{0.50\textwidth}
\caption{\textbf{Ablation on the composition of Rad-CoT.} (b) strips tool calls and observations; (c)--(e) each remove one task type.} 
\label{tab:data}
\end{minipage}

\par\nointerlineskip\vspace{2pt}

\begin{minipage}[t]{0.47\textwidth}
\centering
\tablestyle
\setlength{\tabcolsep}{2.8pt}
\begin{tabular}{@{}llrrr@{}}
\toprule
Benchmark & Base & $n$ & TIR (\%) & Acc.\ (\%) \\
\midrule
\multirow{2}{*}{ChestA.}
& Wrong   & 1123 & \textbf{64.9} & 54.1 \\
& Correct & 1377 & 43.6 & \textbf{86.0} \\
\midrule
\multirow{2}{*}{CheXb.}
& Wrong   & 252  & \textbf{61.5} & 54.4 \\
& Correct & 366  & 45.9 & \textbf{85.5} \\
\bottomrule
\end{tabular}
\end{minipage}%
\hfill
\begin{minipage}[t]{0.50\textwidth}
\centering
\tablestyle
\setlength{\tabcolsep}{3pt}
\begin{tabular}{@{}lccc@{}}
\toprule
Corpus & ChestA. & CheXb. & SLAKE \\
\midrule
\rowcolor{oursgreen}
(a) Full & \textbf{71.7} & \textbf{77.3} & \textbf{93.0} \\
(b) w/o tool traj.  & 67.2 & 71.2 & 86.0 \\
(c) w/o report CoT  & 66.2 & 72.6 & 90.4 \\
(d) w/o MCQ         & 69.8 & 74.9 & 92.1 \\
(e) w/o VQA         & 70.4 & 74.2 & 82.5 \\
\bottomrule
\end{tabular}
\end{minipage}
\end{table}

\textbf{Selective evidence acquisition.}
Tab.~\ref{tab:need_stratified} stratifies questions by the correctness of
direct-answer Qwen3-VL-8B, which we use as a proxy for evidence need.
On ChestAgentBench, X-Ragent invokes auxiliary tools on 64.9\% of questions
missed by the base model, 21.3 percentage points higher than the 43.6\%
invocation rate on questions it already answers correctly (1.49$\times$).
A similar pattern holds on CheXbench, where the invocation rate increases
from 45.9\% on base-correct questions to 61.5\% on base-error questions
(1.34$\times$).
Meanwhile, corrections substantially outnumber regressions, yielding
rescue-to-damage ratios of 3.15$\times$ on ChestAgentBench and
2.58$\times$ on CheXbench.
These results indicate that learned routing concentrates auxiliary acquisition
on cases where the base policy is more likely to fail, rather than invoking
tools uniformly.

\textbf{Data composition.}
Tab.~\ref{tab:data} removes one supervision component at a time while
resampling the remainder to keep the corpus size fixed. Removing tool-trajectory supervision in row (b) lowers accuracy by 4.5, 6.1, and 7.0 points on ChestAgentBench, CheXbench, and SLAKE, respectively.
Removing report-style CoT in row (c) causes the largest drop on the held-out ChestAgentBench (5.5 points), while removing VQA supervision in row (e) mainly affects SLAKE (-10.5 points), reflecting its closed-ended VQA format.

\begin{table}[t]
\centering
\small
\caption{
\textbf{Controlled ablation of acquisition-validity gating on ChestAgentBench.}
The gated and no-gate variants use the same anatomy-measurement backend and
differ only in whether view validity suppresses anatomy evidence.
G-only / NG-only denote paired cases correct only with / without gating.
}
\label{tab:validity_gate}
\setlength{\tabcolsep}{4pt}
\begin{tabular}{@{}lrrrrr@{}}
\toprule
View group & $N$ & No gate & Gated & $\Delta$ & G-only / NG-only \\
\midrule
All
& 2500 & 71.44 & \textbf{71.68} & +0.24 & 17 / 11 \\
Frontal
& 1887 & 72.18 & 72.18 & +0.00 & 8 / 8 \\
Non-frontal / atyp.
& 613 & 69.17 & \textbf{70.15} & +0.98 & 9 / 3 \\
\bottomrule
\vspace{-0.25in} 
\end{tabular}
\end{table}

\begin{figure*}[t]
    \centering
    \includegraphics[width=\linewidth]{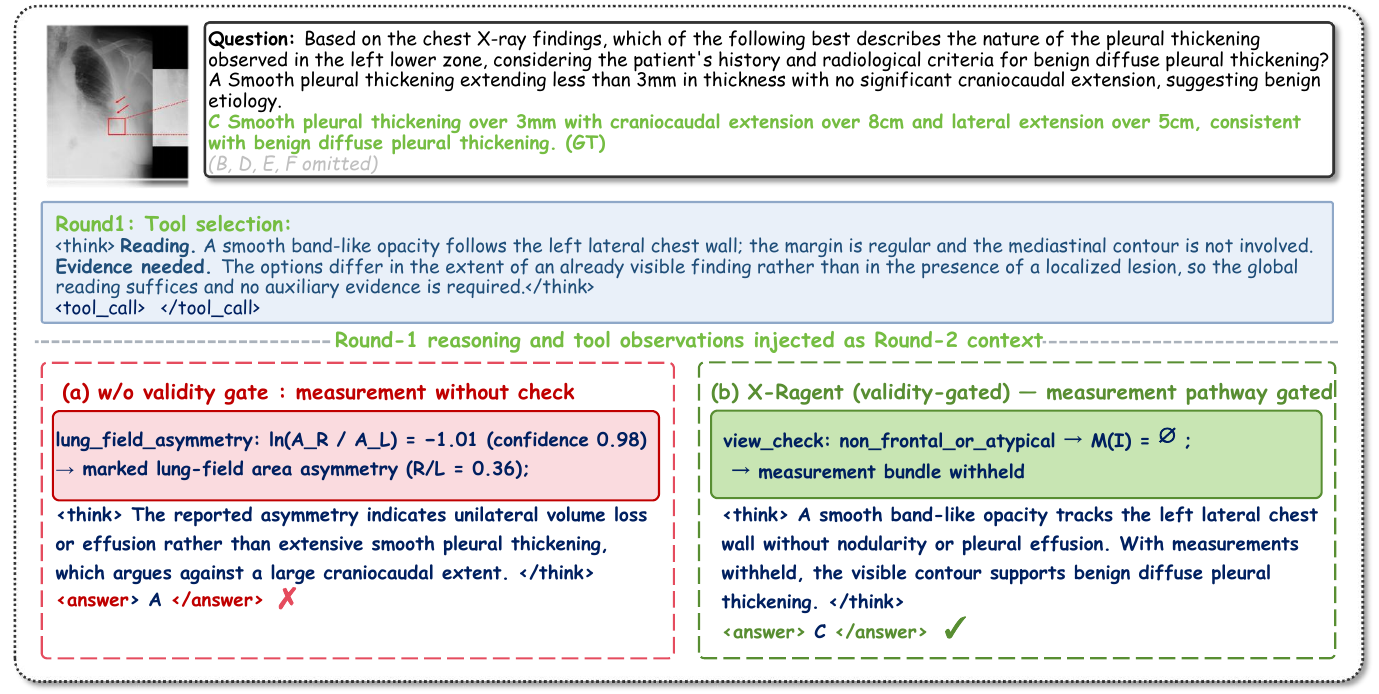}
    \caption{
    \textbf{Counterfactual case study of validity-gated quantitative evidence.}
    Without the validity gate, an inapplicable lung-field asymmetry
    measurement ($\ln(A_R/A_L)=-1.01$) is exposed to the policy and shifts
    the prediction from the correct option C to the incorrect option A.
    X-Ragent instead withholds invalid measurement bundle and returns
    correct answer based on radiographic evidence.
    }
    \label{fig:main_case}
    \vspace{-0.2in}  
\end{figure*}

\textbf{Controlled validity-gate ablation.}
On ChestAgentBench, \texttt{view\_check} prevents measurement execution for 24.5\% of inputs, with 31\% ultimately withheld after the mask-quality check.
In a controlled comparison with the same measurement backend (Tab.~\ref{tab:validity_gate}), validity gating raises accuracy on 613 non-frontal or atypical CXRs from 69.17\% to 70.15\% (+0.98 points), leaves 1,887 frontal CXRs unchanged (72.18\%), and improves overall accuracy from 71.44\% to 71.68\% (+0.24 points).
Paired prediction changes show 8 corrections and 8 regressions on frontal inputs, versus 9 corrections and 3 regressions on non-frontal or atypical inputs, so all six net corrections arise from the validity-sensitive subset.
Fig.~\ref{fig:main_case} provides a representative example: exposing an inapplicable lung-field asymmetry measurement flips the answer from the correct option C to the incorrect option A, whereas withholding it preserves the visually supported answer.
Additional qualitative examples of auxiliary-evidence acquisition and integration are provided in Appendix~\ref{appx:case_study}.

\section{Related Work}

\paragraph{Multimodal Vision--Language Reasoning.}
Multimodal large language models increasingly employ explicit intermediate reasoning. Multimodal-CoT separates rationale generation from answer related inference~\citep{zhang2024multimodalcot}, while LLaVA-CoT structures inference into summarization, visual interpretation, logical reasoning, and conclusion generation~\citep{xu2025llavacot}. VLM-R1 further shows that rule-based reinforcement learning improves visual reasoning and generalization over supervised fine-tuning~\citep{shen2025vlmr1}. However, these methods reason over fixed multimodal inputs rather than acquiring task-specific evidence when needed.

\paragraph{Tool-Augmented Agentic Systems.}
Tool-augmented agents extend static reasoning into an interaction loop involving reasoning, action, and returned observations. ReAct interleaves reasoning with task-specific actions~\citep{yao2023react}, Toolformer learns when and how to invoke external APIs from self-supervised examples~\citep{schick2023toolformer}, and HuggingGPT uses an LLM controller to plan tasks, select specialized models, and integrate their outputs across modalities~\citep{shen2023hugginggpt}. However, these general systems do not address clinical validity, acquisition-dependent applicability, or evidence integration for medical imaging tools.

\paragraph{Medical Vision--Language Reasoning and Agents.}
Medical VLMs have extended image interpretation toward instruction-driven reasoning, as exemplified by CheXagent for chest X-ray understanding~\citep{chen2024chexagent}. Tool-augmented systems further combine general reasoning with specialized medical models: MMedAgent learns to select tools across imaging modalities~\citep{li2024mmedagent}, whereas MedRAX integrates classification, segmentation, and grounding modules within a ReAct-style CXR reasoning loop~\citep{fallahpour2025medrax}. More recent methods such as ViTAR and Ophiuchus use supervised fine-tuning and reinforcement learning to train iterative visual or tool-mediated reasoning~\citep{chen2025vitar,jiang2025ophiuchus}. In contrast, our method learns a compact two-turn CXR interaction: the first turn performs preliminary interpretation and tool planning, while the second integrates returned observations into the final decision. Its reward jointly considers diagnostic correctness, tool cost, and consistency with verifiable radiographic measurements.

\section{Conclusion}
\label{sec:conclusion}
In this work, we propose X-Ragent, a post-training framework that combines learned evidence seeking with validity-gated quantitative augmentation for evidence-grounded CXR reasoning. Trained on structured diagnostic trajectories with agentic SFT and GRPO, X-Ragent learns when to acquire auxiliary evidence and how to integrate it, while admitting quantitative measurements only when clinically valid. X-Ragent achieves state-of-the-art performance across three CXR benchmarks, including an 8.6-point gain on the fully held-out ChestAgentBench and consistent improvements across all seven interpretation capabilities. These results highlight the value of selective evidence acquisition and validity-aware grounding for reliable and generalizable CXR reasoning.

\section*{AI USE STATEMENT}
We clarify the role of Large Language Models (LLMs) in this work. During manuscript preparation, LLMs were used for translation, language polishing and providing pre-submission feedback on the manuscript. In the research pipeline, a locally deployed Qwen3.5-122B-A10B-FP8 was used to generate the synthetic Rad-CoT reasoning trajectories, which were filtered and verified by the authors; a locally deployed Qwen3.5-27B serves as the knowledge source of \texttt{evidence\_lookup}. LLMs were not used to independently formulate research hypotheses, design methodological innovations, or determine scientific conclusions.

\section*{Ethics Statement}
This work uses only de-identified, retrospective data; no new data were collected. MIMIC-CXR-JPG~\citep{PhysioNet-mimic-cxr-jpg-2.1.0} and LLaVA-Rad annotations~\citep{PhysioNet-llava-rad-mimic-cxr-annotation-1.0.0} were accessed under the PhysioNet Credentialed Health Data Use Agreement, and other datasets are used under their respective licenses. Following PhysioNet's guidance on LLM use\footnote{\url{https://physionet.org/news/post/llm-responsible-use/}}, all LLMs in our pipeline were deployed locally, so no MIMIC data were shared with third-party services; any release of Rad-CoT will be made only through PhysioNet. X-Ragent is a research prototype, not a medical device, and should not be used for clinical decision-making.

\bibliography{conference}
\bibliographystyle{conference}

\newpage

\appendix
\section{Appendix}
In this appendix, we provide additional experimental details, tool suite details, extended related work, and the prompts used in our analysis. Detailed contents are as follows:
\setlength{\cftbeforesecskip}{0.5em}
\cftsetindents{section}{0em}{1.8em}
\cftsetindents{subsection}{1em}{2.5em}
\etoctoccontentsline{part}{Appendix}
{
  \etocsettocstyle{}{}
  \localtableofcontents
}

\section{Experiment}
\subsection{More Experimental Details}
\label{appx:exp_details}

\textbf{Datasets.}
ChestAgentBench contains 2,500 six-choice questions. Each question is
associated with four category labels, including a reasoning label shared by
all questions. Following the benchmark protocol, we report seven
non-reasoning capability accuracies: detection (972), classification (1,517),
localization (983), comparison (1,026), relationship (997), diagnosis (1,024),
and characterization (981). These capability subsets overlap at the question
level and are therefore not mutually exclusive.

CheXbench evaluation follows the MedRAX-style axis-2 subset and contains
618 two-choice questions from three sources: 380 OpenI fine-grained reasoning
questions, 115 Rad-ReStruct VQA questions, and 123 SLAKE VQA questions.
The standalone SLAKE evaluation contains 114 closed-ended chest-X-ray
questions and is distinct from the SLAKE subset used within CheXbench.

\textbf{Evaluation metrics.}
For ChestAgentBench, we report accuracy for each of the seven overlapping capability subsets, and report the overall score as question-level accuracy over all 2,500 questions. For CheXbench, we report source-specific accuracy and compute the
overall score as the unweighted macro average across OpenI, Rad-ReStruct,
and SLAKE. For standalone SLAKE, we report exact-match accuracy under the
benchmark's closed-form answer protocol on the 114-question test set.

\textbf{SFT configuration.}
We initialize from Qwen3-VL-8B-Instruct and perform one epoch of supervised
fine-tuning with a learning rate of $1\times10^{-5}$ and an effective batch
size of 16. The maximum sequence length is 8,192 tokens. The vision encoder
is kept frozen during SFT, while the language policy is optimized to learn
both evidence routing and evidence-integrated reasoning.

\paragraph{Hyperparameters.}
Tab.~\ref{tab:hparams} lists the key hyperparameters. Since $R_{\text{CXR}}\ge 0$ and $\lambda_t K_{\max}<1$, any well-formed correct trajectory receives a higher reward than any incorrect one under Eq.~\ref{eq:total_reward}, so tool cost never overrides diagnostic correctness.

\begin{table}[h]
\centering
\caption{\textbf{Key hyperparameters of X-Ragent.}}
\label{tab:hparams}
\small
\begin{tabular}{llc}
\toprule
Stage & Hyperparameter & Value \\
\midrule
\multirow{6}{*}{SFT}
 & Base model & Qwen3-VL-8B-Instruct \\
 & Trainable modules & LLM (vision encoder frozen) \\
 & Epochs / learning rate & 1 / $1\times10^{-5}$ \\
 & Effective batch size & 16 \\
 & Max sequence length & 8,192 \\
 & LR schedule / warmup ratio & cosine / 0.1 \\
\midrule
\multirow{9}{*}{GRPO}
 & Training prompts / steps & 4,715 / 300 \\
 & Learning rate & $2\times10^{-6}$ \\
 & Rollouts per prompt $G$ & 4 \\
 & Rollout / actor update batch size & 16 / 8 \\
 & Max policy-routed calls $K_{\max}$ & 5 \\
 & Clip ratio $\epsilon$ / KL coefficient $\beta$ & 0.2 / 0.01 \\
 & Reward weights $\lambda_m$ / $\lambda_t$ & 0.5 / 0.05 \\
 & Rollout temperature & 1.0 \\
 & Max response length (per turn) & 2,048 \\
\midrule
\multirow{2}{*}{Inference}
 & Decoding & greedy \\
 & Max new tokens (per turn) & 2,048 \\
\bottomrule
\end{tabular}
\end{table}

\subsection{Rad-CoT Construction and Quality Control}
\label{app:data_construction}

\begin{figure}[t]
    \centering
    \includegraphics[width=\linewidth]{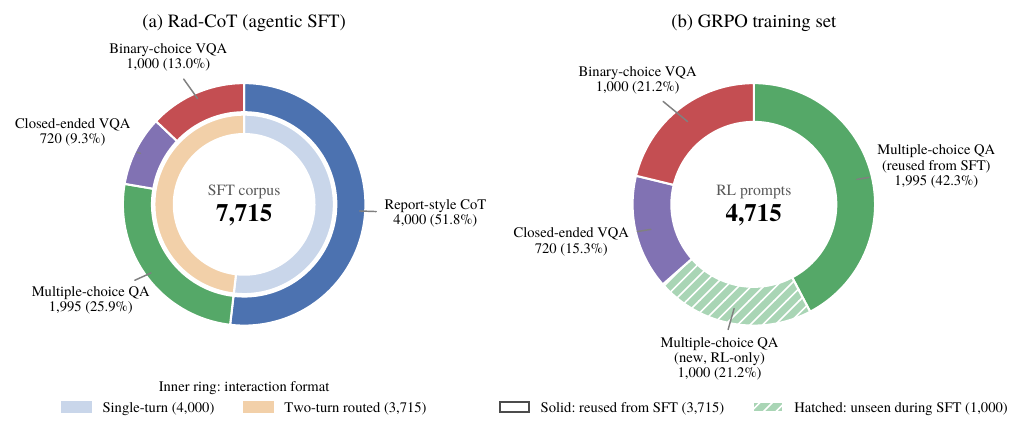}
    \caption{\textbf{Composition of the training data.}
    (a) The SFT corpus Rad-CoT. The outer ring shows the task type; the inner
    ring distinguishes single-turn report-style CoT from two-turn routed
    examples, in which the first-turn tool plan may be empty.
    (b) The GRPO corpus. Solid segments are reused from the routed SFT
    examples, while the hatched segment denotes multiple-choice questions
    unseen during SFT.}
    \label{fig:data_dist}
\end{figure}

\textbf{Corpus composition.}
Rad-CoT contains 7,715 SFT instances, of which 3,715 follow the two-turn
routed protocol and 4,000 are single-turn report-style CoT examples.
The GRPO corpus contains 4,715 examples: all 3,715 routed SFT examples are
reused, and 1,000 additional multiple-choice questions unseen during SFT
are added. Report-style CoT is used only for SFT, since it lacks a
rule-verifiable answer. Fig.~\ref{fig:data_dist} visualizes the composition
of both corpora, and Tab.~\ref{tab:radcot_composition} lists the exact counts
together with the interaction format and reward type of each subset.
We use ``7.7K'' and ``4.7K'' in the main text as rounded descriptions of the
two corpora.

\begin{table}[t]
\centering
\small
\caption{\textbf{Subset statistics of the SFT and GRPO corpora.}
``Routed'' denotes the two-turn protocol in Sec.~\ref{sec:Problem}, where the
first-turn tool plan may be empty. GRPO uses only subsets with rule-verifiable
answers; all GRPO rewards are binary.}
\label{tab:radcot_composition}
\setlength{\tabcolsep}{4pt}
\begin{tabular}{@{}lllrr@{}}
\toprule
Subset & Format & Answer / reward & SFT & GRPO \\
\midrule
Report-style diagnostic CoT   & Single-turn & Free-text findings        & 4,000 & -- \\
Multiple-choice QA            & Routed      & Option-letter match       & 1,995 & 1,995 \\
Multiple-choice QA (RL-only)  & Routed      & Option-letter match       & --    & 1,000 \\
Closed-ended VQA              & Routed      & Normalized exact match    & 720   & 720 \\
Binary-choice VQA             & Routed      & Option-letter match       & 1,000 & 1,000 \\
\midrule
\textbf{Total}                &             &                           & \textbf{7,715} & \textbf{4,715} \\
\bottomrule
\end{tabular}
\end{table}

\textbf{Report-style reasoning distillation.}
The report-style subset is constructed from PA radiographs in MIMIC-CXR-JPG
paired with reference findings derived from the LLaVA-Rad annotations. We
first stratify 6,000 candidate studies and use locally deployed Qwen3.5-122B-A10B-FP8 to synthesize the
intermediate diagnostic rationale conditioned on both the radiograph and its
reference findings. Importantly, the teacher is used to generate the reasoning
process rather than to freely regenerate the diagnostic target: after
generation, the reference findings are retained as the target answer. This
answer-conditioned construction reduces target drift while providing explicit
supervision for mapping radiographic evidence to diagnostic conclusions.

\textbf{Reasoning-quality filtering.}
We apply deterministic quality control to the generated report-style
trajectories. Before generation, we retain PA-view studies with an available
image and non-trivial reference findings. After generation, we remove prompt
or metadata leakage, reject malformed or abnormal-length reasoning, and
discard rationales that excessively reproduce the reference findings. In
particular, candidate rationales are required to contain 40--900 words and
reasoning--finding similarity above 0.82 is rejected. When multiple reasoning
segments are produced, we retain the longest valid segment. Failed generations
are retried once and discarded if they remain invalid. This filtering reduces
the initial 6,000 candidates to 4,000 report-style reasoning trajectories.

\paragraph{Leakage-controlled MCQ construction and source disjointness.}
All 2,995 training MCQs (1,995 for SFT and 1,000 RL-only) are generated from
MIMIC-CXR-JPG PA radiographs~\citep{PhysioNet-mimic-cxr-jpg-2.1.0}. We remove
report-derived cues unavailable at inference, such as sentences referring to
hidden findings (e.g., ``according to the report''), and strip findings
metadata from the final records, so that answers must rely on the radiograph
and query. ChestAgentBench~\citep{fallahpour2025medrax} is instead built from
Eurorad cases, sharing no images, cases, or reports with our training data,
and none of its questions or images is used during MCQ generation.

\textbf{Closed-ended VQA construction.}
We include 720 chest-X-ray closed-ended VQA trajectories to preserve concise
answering behavior. Images overlapping with the SLAKE evaluation split are
removed during construction, preventing direct image overlap between this
training subset and the corresponding held-out evaluation data.

\textbf{Causally ordered tool trajectories.}
The 3,715 routed SFT examples, consisting of 1,995 MCQ, 720 closed-ended VQA, and 1,000 binary-choice VQA examples, follow the two-turn interaction protocol. In the first turn, the policy observes the radiograph and
clinical query and produces an evidence-acquisition decision. Only after this
decision is generated are the selected tool observations inserted into the
conversation. The second assistant turn then integrates the returned evidence
with the original image and query to produce the final reasoning and answer.
Thus, tool observations never precede the actions that request them, matching
the causal interaction structure used at inference time.

During SFT, loss is applied only to policy-generated assistant tokens, whereas
externally injected tool observations are treated as context and masked from
the language-modeling objective. This prevents the policy from being trained
to imitate tool outputs and instead teaches it to decide when to acquire
evidence and how to use the returned evidence.

\textbf{Distribution and integrity checks.}
Finally, we use CheXbert to inspect the diagnostic-label distribution of the
report-style subset and verify coverage across the 14 standard CXR categories.
Together with leakage removal, evaluation-image deduplication, and
generation-time filtering, these checks are intended to ensure that gains from
Rad-CoT reflect learned radiographic reasoning and evidence integration rather
than target leakage or duplicated evaluation content.

\textbf{Evaluation-split integrity.}
For the CheXbench axis-2 evaluation, we construct the exclusion set from
the images actually used by the evaluated subsets rather than excluding the
entire CheXbench image index. The OpenI fine-grained reasoning subset contains
363 unique image references, while the Rad-ReStruct evaluation contributes
94 OpenI-backed image references; their union contains 452 unique evaluation
image references. We exclude this evaluation-specific image set from training
and additionally check for OpenI report-level leakage through sibling images. For the final training set, both image-level overlap with this
452-image exclusion set and OpenI report-level sibling-image leakage are zero.

\subsection{Effect of Prompt-Elicited Reasoning}
\label{app:prompt_reasoning}

We examine whether explicit diagnostic reasoning can be reliably elicited
from the Qwen3-VL-8B base model through prompting alone, without any
post-training. We compare four reasoning-oriented prompts (v0--v3), all of
which explicitly require a \texttt{<think>...</think>} reasoning trace,
against a direct-answer prompt (v4) that does not require an explicit
reasoning block. All settings use the same base checkpoint, image--question
inputs, decoding configuration, and evaluation protocol; only the system
prompt is changed.

For consistency with the main experiments, ChestAgentBench accuracy is
reported as question-level accuracy over all 2,500 questions,
CheXbench accuracy as the unweighted mean across its three data sources,
and SLAKE using closed-ended exact-match accuracy.

\begin{table}[t]
\centering
\small
\caption{\textbf{Effect of prompt-elicited reasoning on the Qwen3-VL-8B base model.}
We compare a direct-answer prompt with four variants that explicitly require
intermediate reasoning. Accuracy (\%) is reported using the same benchmark-level
aggregation protocol as in the main experiments.}
\label{tab:prompt_reasoning}
\setlength{\tabcolsep}{4pt}
\begin{tabular}{lcccc}
\toprule
Prompting Strategy & ChestAgent & CheXbench & SLAKE & Avg. \\
\midrule
\rowcolor{oursgreen}
\textbf{Direct answering (v4)} & \textbf{55.1} & \textbf{65.5} & \textbf{72.8} & \textbf{64.5} \\
Reasoning v0 & 47.6 & 56.7 & 59.6 & 54.6 \\
Reasoning v1 & 47.3 & 56.7 & 45.6 & 49.9 \\
Reasoning v2 & 51.1 & 57.3 & 62.3 & 56.9 \\
Reasoning v3 & 50.3 & 60.1 & 59.6 & 56.7 \\
\bottomrule
\end{tabular}
\end{table}

As shown in Tab.~\ref{tab:prompt_reasoning}, explicitly requiring a reasoning
trace does not improve the base model. Direct answering achieves 55.1\%,
65.5\%, and 72.8\% on ChestAgentBench, CheXbench, and SLAKE, respectively.
Even the strongest explicit-reasoning variant on each benchmark remains lower
by 4.0, 5.4, and 10.5 percentage points, respectively.

These results indicate that simply eliciting longer or more explicit reasoning
from the base model is insufficient to establish reliable diagnostic reasoning
behavior. Together with the RL-only result in Tab.~\ref{tab:pipeline}, where
direct GRPO from the base policy remains substantially weaker than supervised
initialization, this motivates our two-stage post-training strategy. We therefore
first use Rad-CoT to initialize structured diagnostic reasoning and tool-use
behaviors through SFT, and subsequently refine the policy with GRPO.

\subsection{Reward Details}\label{app:reward}

\textbf{Radiographic evidence consistency.} For measurement $k$, let $r_k\in\{0,1\}$ indicate query relevance under a fixed measurement--task correspondence and $b_k\in\{0,1\}$ indicate whether the measurement is exposed by the validity-gated backend. We define its reward eligibility as $a_k=r_kb_k$, where query relevance affects reward eligibility only and does not control measurement execution. The evidence-consistency reward is $R_{\mathrm{CXR}}(\tau)=\frac{\sum_k a_k s_k}{\max(1,\sum_k a_k)}$, where $s_k$ is a rule-based score indicating whether the trajectory interprets measurement $k$ consistently with its verified value. Each measurement contributes at most once.

\textbf{Auxiliary tool-use cost.} $R_{\mathrm{tool}}$ counts policy-generated auxiliary calls from $\mathcal{T}_{\pi}$, with repeated calls counted separately. Backend validity checks and deterministic measurement routines are excluded from this cost.

\subsection{More Ablation Studies}
\label{appx:ablation_study}

\begin{table}[t]
\centering
\small
\caption{\textbf{Ablation on evidence-acquisition modules at inference.}
A single trained checkpoint is evaluated with individual modules selectively
disabled. The validity-gated measurement backend is retained unless objective
measurement is explicitly removed; row~(f) statically invokes all
policy-callable auxiliary tools on every question.}
\label{tab:toolkit}
\begin{tabular}{@{}llc@{}}
\toprule
& Evidence acquisition & ChestAgent \\
\midrule
\rowcolor{oursgreen}
(a) & \textbf{Full (learned routing)} & \textbf{71.7} \\
(b) & w/o visual enh. & 70.1 \\
(c) & w/o aux.\ interp. & 69.5 \\
(d) & w/o obj.\ meas. & 70.8 \\
(e) & w/o all evidence modules & 67.4 \\
(f) & All auxiliary tools (static) & 71.0 \\
\bottomrule
\end{tabular}
\end{table}

\textbf{Module contribution.}
Tab.~\ref{tab:toolkit} disables individual evidence-acquisition modules at
inference using a single trained checkpoint; row~(e) is equivalent to
row~(d) of Tab.~\ref{tab:pipeline}. With all modules removed, accuracy drops
by 4.3 points to 67.4\%, indicating that part of the gain is internalized
during post-training. Auxiliary interpretation contributes most (2.2 points),
followed by visual enhancement (1.6) and objective measurement (0.9), whose
narrower scope and 31\% withholding rate limit its effect. Forcing all
auxiliary tools on every question in row~(f) remains 0.7 points below learned
routing, which invokes tools on only 53.2\% of questions
(Tab.~\ref{tab:need_stratified}), showing that selective routing loses no
accuracy while avoiding unnecessary calls.

\begin{table}[t]
\centering
\small
\caption{\textbf{Ablation on quantitative measurement routing.}
We compare the default X-Ragent design, where quantitative measurements are supplied
by an independent validity-gated backend, with a variant that additionally
requires the first-stage policy to select the measurement tool before validity
checking. Accuracy (\%) is reported on ChestAgentBench and CheXbench.}
\label{tab:measurement_routing}
\setlength{\tabcolsep}{6pt}
\begin{tabular}{@{}lcc@{}}
\toprule
Measurement pathway & ChestAgent & CheXbench \\
\midrule
\rowcolor{oursgreen}
\textbf{Independent validity gate (X-Ragent)} & \textbf{71.7} & \textbf{77.3} \\
Policy routing + validity gate            & 71.0          & 76.1          \\
\bottomrule
\end{tabular}
\end{table}

\textbf{Quantitative measurement routing.}
We further examine whether quantitative measurements should be routed by the
first-stage policy. In the default X-Ragent design, quantitative measurement is
decoupled from policy-routed auxiliary evidence: the validity-gated backend is
applied independently to each input, and measurements are exposed whenever
their acquisition conditions are satisfied. We compare this design with a
variant in which the measurement tool is included in the first-stage action
space and is executed only when explicitly selected by the policy.

As shown in Tab.~\ref{tab:measurement_routing}, the default design improves
accuracy from 71.0\% to 71.7\% on ChestAgentBench and from 76.1\% to 77.3\%
on CheXbench. The policy-routed variant is close to removing measurement
entirely (70.8\%, Tab.~\ref{tab:toolkit}(d)), suggesting that routing tends to
omit these cues even when informative. Since cardiac size, lung-field
asymmetry, and thoracic shift are cheap, verifiable, and complementary to the
MLLM's global visual representation, we supply them directly whenever
acquisition conditions are satisfied, while the policy selectively acquires
task-specific auxiliary evidence.

\subsection{More Case Studies}
\label{appx:case_study}
\begin{figure*}[t]
    \centering
    \includegraphics[width=\linewidth]{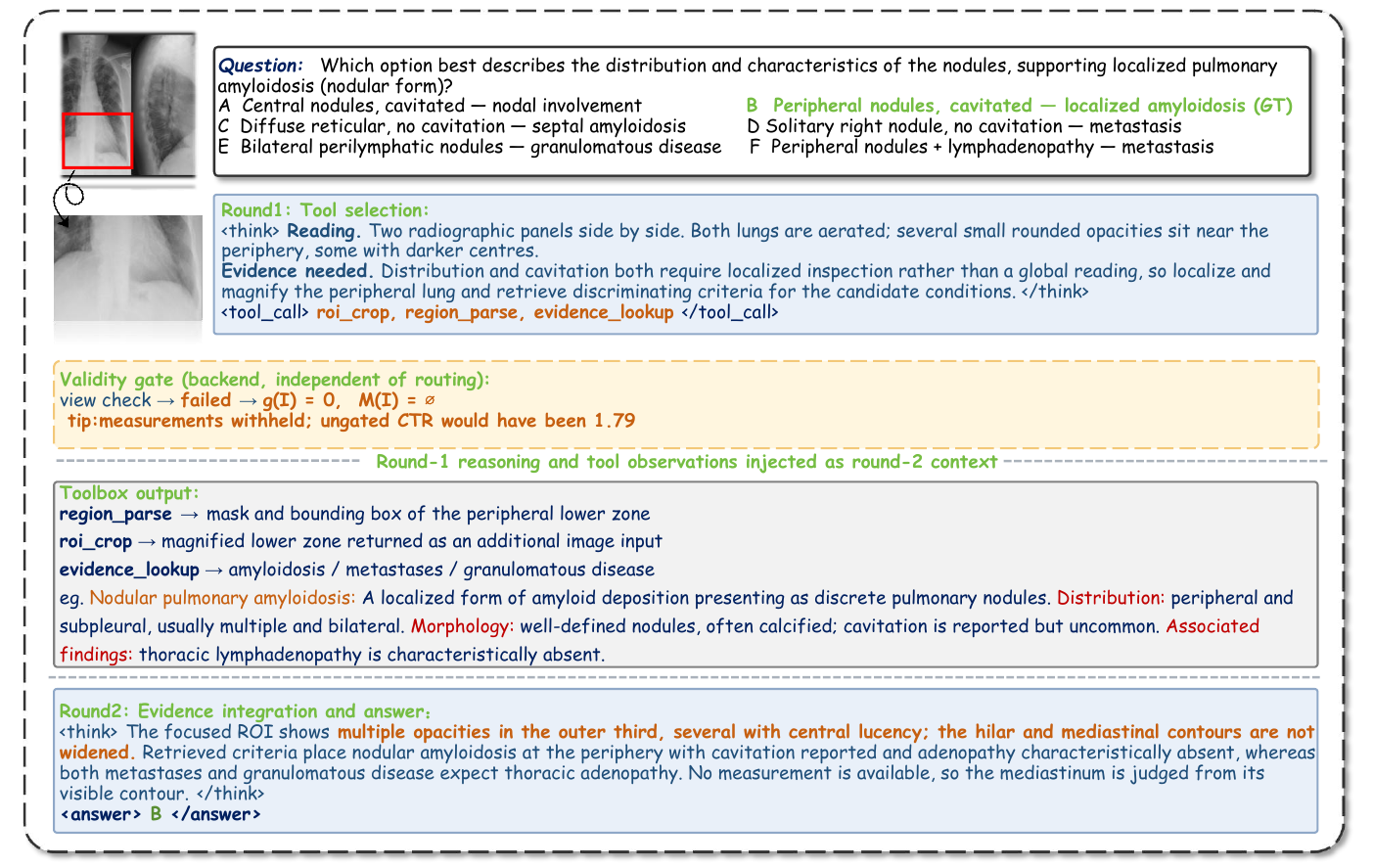}
    \caption{
    \textbf{Selective auxiliary-evidence acquisition and integration.}
    In the first turn, X-Ragent judges global inspection insufficient and
    routes localized visual refinement (\texttt{region\_parse} $\rightarrow$
    \texttt{roi\_crop}) together with condition-specific
    \texttt{evidence\_lookup}. Independently of this routing decision, the
    validity gate rejects the atypical input and withholds the measurement
    bundle ($M(I)=\varnothing$), preventing a physically implausible CTR
    (1.79 if ungated) from entering the reasoning. In the second turn, the
    policy integrates the magnified ROI with the retrieved criteria,
    identifying peripheral cavitated nodules without thoracic adenopathy,
    and selects the correct answer, localized nodular pulmonary amyloidosis.
    }
    \label{fig:aux_case}
\end{figure*}

\textbf{Selective acquisition and evidence integration.}
Fig.~\ref{fig:aux_case} provides an additional example of the learned
auxiliary-evidence pathway. Because the answer options differ in nodule
distribution and cavitation, the first-turn policy determines that localized
inspection is needed and requests a focused ROI via \texttt{region\_parse}
and \texttt{roi\_crop}, together with condition-specific criteria via
\texttt{evidence\_lookup}. In parallel, the validity gate, which runs
independently of the policy's routing decision, rejects the input and
withholds the quantitative measurement bundle; had it been exposed, the
ungated CTR would have been 1.79, a physically implausible value. The second
turn then integrates the localized visual findings with the retrieved
radiographic criteria. The absent measurement is treated as unavailable
rather than normal, and the mediastinum is assessed from its visible
contour. The policy correctly identifies localized nodular pulmonary
amyloidosis.

\section{Tool Suite Details}
\label{appx:tool_library}

X-Ragent separates evidence acquisition into two operationally distinct
pathways. First, the policy selectively invokes auxiliary tools from
$\mathcal{T}_{\pi}
=
\{\texttt{roi\_crop},\texttt{region\_parse},\texttt{evidence\_lookup}\}$
to acquire additional visual or interpretive evidence. These actions are
generated by the policy and are therefore counted in
$R_{\mathrm{tool}}$. Second, quantitative radiographic measurements are
computed by an independent backend whose execution is controlled by an
acquisition-validity check. Backend validity checks and deterministic
measurement routines are not policy actions and are excluded from
$R_{\mathrm{tool}}$.

\begin{table}[t]
\centering
\small
\caption{\textbf{Summary of the X-Ragent evidence-acquisition modules.}
Policy-routed tools are selected by the learned policy, whereas the
validity check and quantitative measurements are executed automatically
by the backend.}
\label{tab:tool_summary}
\setlength{\tabcolsep}{4pt}
\begin{tabularx}{\linewidth}{@{}l l l X@{}}
\toprule
Component & Execution & Backbone & Returned evidence \\
\midrule
\texttt{region\_parse}
& Policy-routed
& CXAS / BiomedParse
& Region mask and bounding box \\

\texttt{roi\_crop}
& Policy-routed
& Geometric crop
& Local high-resolution view \\

\texttt{evidence\_lookup}
& Policy-routed
& Qwen3.5-27B (local API)
& Condition-specific radiographic knowledge \\

\texttt{view\_check}
& Backend
& OpenCXR
& Acquisition applicability \\

Measurement bundle
& Backend
& CXAS + deterministic rules
& CTR, lung asymmetry, thoracic shift \\
\bottomrule
\end{tabularx}
\end{table}

\subsection{Visual Refinement Tools}

\paragraph{Region Parsing.}
\texttt{region\_parse} provides localized anatomical or pathological
evidence for region-specific CXR inspection. Following the specialized
parsing tools used in prior medical agents
\citep{jiang2025ophiuchus}, chest anatomical
structures are parsed using CXAS \citep{seibold2023cxas}, while
text-specified pathological regions can be localized using
BiomedParse \citep{zhao2024biomedparse}. Given the input radiograph and
a target structure or finding, the parser returns a segmentation mask
together with its tight axis-aligned bounding box. These outputs provide
explicit region-level grounding and can subsequently be used for
fine-grained visual inspection.

\paragraph{ROI Crop.}
\texttt{roi\_crop} is a deterministic visual-refinement operator that
magnifies a region identified by \texttt{region\_parse}. Specifically,
given the returned bounding box
$b=(x_1,y_1,x_2,y_2)$, the corresponding region is cropped from the
original radiograph,
\begin{equation}
I_{\mathrm{crop}}
=
I[y_1:y_2,\;x_1:x_2],
\end{equation}
and returned as an additional visual observation. The original CXR is
retained in the multimodal context, allowing the policy to jointly
reason over the global image and the localized high-resolution view.
This parse--then--crop design follows the general segment/localize--then--zoom
strategy used in tool-augmented visual reasoning
\citep{jiang2025ophiuchus}.

\subsection{Auxiliary Interpretation via Evidence Lookup}

To mitigate cross-modal hallucination, \texttt{evidence\_lookup} queries a locally deployed Qwen3.5-27B through an OpenAI-compatible API, invoked online (i.e., on demand during inference). Given the candidate conditions specified by the policy, it returns a
concise description of their characteristic CXR manifestations, including
typical appearance, distribution, and discriminating imaging features. This
tool supplies general condition-level knowledge rather than a patient-specific
diagnosis; the returned description is appended to the query as $E_C$ and
reconciled with the visual findings by the policy in the second turn.

\subsection{Validity-Gated Objective Measurements}

\paragraph{Validity Gate.}
Unlike the auxiliary tools above, \texttt{view\_check} is executed
automatically and independently of the policy routing decision. We use
the Image Sorter from OpenCXR%
\footnote{\url{https://github.com/DIAGNijmegen/opencxr}},
which distinguishes frontal PA, frontal AP, lateral, and non-CXR
images. Because the quantitative measurements used in X-Ragent are defined
for standard PA radiographs, the measurement pathway is enabled only
when the image satisfies the required PA acquisition condition.
Otherwise, the quantitative measurement bundle is withheld.

We additionally suppress the bundle when the anatomical masks required
for measurement are unavailable or fail the backend quality checks.
Thus, the validity gate can be written as
\begin{equation}
g(I)
=
g_{\mathrm{view}}(I)\,
g_{\mathrm{quality}}(I),
\end{equation}
and quantitative evidence is exposed only when $g(I)=1$.
Neither \texttt{view\_check} nor the subsequent measurement routines
contributes to $R_{\mathrm{tool}}$.

\paragraph{Measurement Bundle.}
For an applicable input, the backend uses CXAS-derived anatomical masks
to compute three deterministic radiographic measurements:
cardiothoracic ratio (CTR), lung-field asymmetry, and thoracic shift.

\textbf{Cardiothoracic ratio.}
Following mask-based CTR estimation~\citep{chamveha2020automated}, we identify the image row where
the heart mask attains its maximum horizontal width. Let $w_c$ denote
the cardiac width on this row and $w_t$ the horizontal span of the
bilateral lung fields at the same level. The cardiothoracic ratio is
\begin{equation}
\mathrm{CTR}
=
\frac{w_c}{w_t}.
\end{equation}
On a valid PA radiograph, $\mathrm{CTR}>0.5$ is interpreted as an
enlarged cardiac silhouette.

\textbf{Lung-field asymmetry.}
Let $A_L$ and $A_R$ denote the projected areas of the left and right lung masks, respectively. 
We compute a log-ratio bilateral asymmetry
index as
\begin{equation}
\mathrm{LFA}=\ln\frac{A_R}{A_L},
\end{equation}
where the magnitude reflects the degree of lung-field asymmetry and the
sign indicates its direction.

\textbf{Thoracic shift.}
Thoracic shift is estimated geometrically from CXAS-derived central
thoracic structures relative to the bilateral thoracic midline. The
result records both the magnitude and direction of displacement and is
used as objective evidence for queries involving mediastinal or
tracheal displacement.

The resulting measurements are converted to fixed categorical
interpretations by deterministic rules and returned to the policy as a
structured observation. These verified interpretations are also used
to compute the evidence-consistency score $s_k$ in
$R_{\mathrm{CXR}}$ (Appendix~\ref{app:reward}).

If the applicability check fails, we define
\begin{equation}
M(I)=\varnothing,
\end{equation}
so that no potentially unreliable quantitative value enters the
reasoning context. Importantly, an absent measurement therefore denotes
\emph{unavailable or inapplicable evidence}, rather than a normal
radiographic finding.

\subsection{Tool Execution and Observation Injection}

For each example, policy-routed tools are executed only after their
corresponding actions have been generated. Returned crops, parsing
results, and knowledge-lookup observations are then injected into the
second-turn context together with any valid quantitative measurements.
During SFT, externally generated observation spans are masked from the
language-modeling loss, such that the policy is optimized to learn tool
selection and evidence-integrated reasoning rather than to reproduce
the outputs of the external tools.

The same tool implementations and deterministic measurement routines
are used during Rad-CoT construction and inference, and quantitative
measurements are precomputed and cached whenever possible to ensure
consistent observations across training and evaluation.

\section{Detailed Related Work}
\label{appx:detailed_rw}

This section extends the main-text related work to broader directions connected to X-Ragent.

\paragraph{Post-Training and Reasoning in Foundation Models.}
Reinforcement learning with verifiable rewards (RLVR) has become a standard route for strengthening reasoning, and recent work extracts more learning signal from each rollout group.
EchoRL~\citep{bi2026echorl} recovers signal from all-success groups by feeding back an entropy-selected clip of a successful rollout as auxiliary supervision, ReflectRL~\citep{bi2026reflectrllearninggoldennegative} uses failed expert trajectories as material for reflection rather than imitation, and RGR-GRPO~\citep{bi2025reward} provides dense rubric-based rewards and off-policy guidance for multi-domain reasoning.
Other work assesses or steers reasoning without additional training: GeoR~\citep{bi2026the} judges correctness from the layerwise evolution of hidden states, PromptCD~\citep{bi2026promptcd} contrasts responses under paired positive and negative prompts, and ADE-CoT~\citep{qu2026scale} makes test-time scaling for image editing efficient through difficulty-aware budgets and early stopping.
On the data side, RefineX~\citep{bi2025refinex} distills expert refinements into compact deletion programs for cleaning pre-training corpora, LLMSurgeon~\citep{luo2026llmsurgeon} infers the domain mixture of a pre-training corpus from generated text, PRISM~\citep{Bi2025PRISMSI} selects visual instruction data without training, and DenseFusion-1M~\citep{li2024densefusion} merges multiple vision experts into dense captions.
In medicine, EMPOWER~\citep{chen2026empower} evolves clinical prompts under a multi-dimensional quality assessment.
X-Ragent likewise relies on verifiable rewards and data filtering, but its accuracy gate restricts process-level signals to correct trajectories.

\paragraph{Tool-Augmented Agents and Process Rewards.}
A growing line of work trains multimodal agents to decide when to invoke external tools.
Video-STAR~\citep{yuan2025video} learns tool use for open-vocabulary action recognition under a hierarchical reward that accounts for tool efficiency, IndusAgent~\citep{tan2026indusagent} uses a gated reinforcement learning objective so that cropping, enhancement, and prior retrieval are invoked only when beneficial, and FactGuard~\citep{li2026factguard} selectively seeks external evidence for video misinformation detection through agentic fine-tuning and decision-aware reinforcement learning.
ImagineAgent~\citep{yuan2026if} adds a generative model that synthesizes alternative viewpoints to resolve occlusion in human--object interaction recognition, while AutoDesign~\citep{luo2026autodesign} recursively improves the agent harness rather than model weights for long-horizon poster design.
Process-level supervision is also emerging: CaC~\citep{wang2026cac} trains video reward models with two-turn spatiotemporal grounding under GRPO and IoU rewards, and RL3DEdit~\citep{wang2026geometry} rewards multi-view consistency with signals from a 3D foundation model.
X-Ragent shares this selective tool use while separating policy-routed tools from deterministic measurements admitted by a clinical validity gate.

\paragraph{Multimodal Understanding, Generation, and Retrieval.}
Strong visual representations and efficient multimodal backbones underpin such agents.
mc-BEiT~\citep{li2022mc} refines masked image modeling with soft multi-choice visual-token targets, random-label bridge training~\citep{luo2026language} aligns language-pretrained parameters to vision tasks without manual labels, and $\gamma$-MoD~\citep{luo2025gamma} converts dense MLLM layers into mixture-of-depth layers to reduce training and inference cost.
LongCat-Next~\citep{team2026longcat} represents text, vision, and audio in a shared discrete token space, while SIRI-Bench~\citep{song2025siri} shows that current VLMs struggle with spatial reasoning over 3D scenes.
InstructHOI~\citep{luo2026instructhoi} fuses visual interaction cues with context-aware instructions for human--object interaction detection, and synergistic prompting~\citep{luo2025synergistic} lets visual and textual prompts mutually refine each other.
Composed retrieval poses a related alignment problem: ConeSep~\citep{ConeSep} unlearns noisy triplet correspondences, TEMA~\citep{TEMA} maps textual clauses to image entities for multi-modification queries, and ReTrack~\citep{ReTrack} calibrates composed features biased toward the reference video.
In remote sensing, PromptCC~\citep{liu2023decoupling} decouples change captioning into detection and description, and a recent survey~\citep{liu2025remoteVLM} traces spatiotemporal vision--language models toward LLM-based foundation models.
These works underline the difficulty of aligning fine-grained visual evidence with language, which X-Ragent addresses by injecting localized views and condition-specific knowledge.

\paragraph{Vision--Language--Action Models and Embodied Reasoning.}
VLA models extend multimodal reasoning to physical decision making; a recent survey~\citep{zhang2025pure} organizes them into autoregressive, diffusion-based, reinforcement-based, and hybrid paradigms.
In autonomous driving, CoC-VLA~\citep{NEURIPS2025_66d09284} transfers long-tail skills from simulation via chain-of-causality reasoning and adversarial training, Reasoning-VLA~\citep{zhangreasoning} decodes trajectories in parallel through learnable action queries, AutoDrive-R$^2$~\citep{yuan2025autodrive} pairs self-reflective chain-of-thought data with GRPO under physics-grounded rewards, and MapExpert~\citep{zhang2025mapexpert} routes map elements to sparse experts for online HD map construction.
Pelican-Unify~\citep{zhang2026pelican} jointly predicts future video and actions from a shared latent, and Pelican-VLA~\citep{ding2026pelican} shows that a compact reasoning bottleneck induces attention to task-relevant regions before fine-tuning.
In robotics, VGA~\citep{song2026robotic} treats manipulation as vision-to-geometry mapping with a pretrained 3D geometry model, and CMAST~\citep{song2026human} discovers human-assistive tasks through multi-agent reasoning over a scalable search tree.
Like the physics-grounded rewards of driving VLAs, X-Ragent grounds decisions in verifiable radiographic measurements rather than textual priors alone.

\paragraph{Geometry-Aware Perception and 3D Scene Modeling.}
Explicit structural priors have repeatedly improved dense geometric inference.
In multi-view stereo, a series of methods constrains patch deformation with segmentation and edge priors, including SAM-based instances in SD-MVS~\citep{yuan2024sd}, textureless-region planarization in TSAR-MVS~\citep{yuan2024tsar}, multi-granularity depth edges in MSP-MVS~\citep{yuan2025msp}, visibility-aware deformation in DVP-MVS~\citep{yuan2025dvp} and DVP-MVS++~\citep{yuan2025dvpplusplus}, and panoptic-segmentation edges in SED-MVS~\citep{yuan2025sed}.
Learning-based methods inject monocular foundation-model features (MonoMVSNet~\citep{monomvsnet}) or efficient Mamba-based global aggregation (MVSMamba~\citep{mvsmamba}), and FE2E~\citep{wang2026edit} shows that image-editing diffusion models are better initializations than text-to-image generators for depth and normal estimation.
For 3D scenes, Layout2Scene~\citep{chen2025layout2scene} generates scenes from semantic layouts with diffusion priors, STDR~\citep{li2025stdr} disentangles spatial and temporal patterns of Gaussian primitives, and GradiSeg~\citep{li2024gradiseg} refines Gaussians along object boundaries for 3D segmentation.
Analogous to the anatomical masks behind our measurement backend, these methods show that segmentation and geometric priors make dense estimates more trustworthy.

\paragraph{Transferability, Generalization, and Optimization.}
Reliable learning systems must also transfer beyond their training conditions.
PED~\citep{li2023exploring} estimates the transferability of pre-trained models without fine-tuning, DSU~\citep{li2022uncertainty} models feature statistics as Gaussian random variables to simulate domain shifts, ZeroDiff~\citep{pmlr-v306-fan26d} reconstructs time series at unmeasured locations with a diffusion-calibrated prior, and simulated lifting~\citep{shen2026escaping} escapes spurious local minima in matrix sensing by emulating over-parameterization.
These studies reflect a broader trend of injecting informative priors into learning systems, which X-Ragent pursues through validity-gated quantitative evidence.

\section{Prompts}
\label{appx:prompts}

\subsection{Prompts for Base-Model Reasoning Analysis}
\label{app:reasoning_prompts}

For the prompting analysis in Appendix~\ref{app:prompt_reasoning}, we evaluate
Qwen3-VL-8B-Instruct using one direct-answer prompt and four
reasoning-oriented variants. The image, question, answer candidates, decoding
configuration, and evaluation protocol are kept identical across all settings.

\paragraph{Direct Answering (v4).}
\begin{quote}
\small
You are a radiologist specialized in chest X-ray interpretation. Analyze the
image and the question, then give the final answer inside
\texttt{<answer>...</answer>}. If the question provides lettered options, put a
single option letter in the answer tag. If it is a yes/no question, put exactly
`Yes' or `No'. Otherwise put the shortest clinically accurate answer.
\end{quote}

\paragraph{Generic Step-by-Step Reasoning (v0).}
\begin{quote}
\small
You are an expert radiologist. Reason step by step inside
\texttt{<think>...</think>}, then give the final answer inside
\texttt{<answer>...</answer>}. If the question provides lettered options, put a
single option letter in the answer tag. If it is a yes/no question, put exactly
`Yes' or `No'. Otherwise put the shortest clinically accurate answer.
\end{quote}

\paragraph{CXR-Specific Step-by-Step Reasoning (v1).}
\begin{quote}
\small
You are a radiologist specialized in chest X-ray interpretation. For every
question, first reason step by step inside \texttt{<think>...</think>} about the
visual findings and the question. Then give your final answer inside
\texttt{<answer>...</answer>}: a single option letter if choices are provided,
exactly `Yes' or `No' for yes/no questions, or the shortest accurate phrase
otherwise.
\end{quote}

\paragraph{Findings-First Reasoning (v2).}
\begin{quote}
\small
You are a radiologist specialized in chest X-ray interpretation. Inside
\texttt{<think>...</think>}, first describe the key imaging findings, then reason
toward the answer. After the closing tag, give the final answer inside
\texttt{<answer>...</answer>}: a single option letter if choices are provided,
exactly `Yes' or `No' for yes/no questions, or the shortest accurate phrase
otherwise.
\end{quote}

\paragraph{Concise Reasoning (v3).}
\begin{quote}
\small
You are a medical imaging expert. Reason briefly inside
\texttt{<think>...</think>} (a few sentences), then give the final answer inside
\texttt{<answer>...</answer>}: a single option letter if choices are provided,
exactly `Yes' or `No' for yes/no questions, or the shortest accurate phrase
otherwise. Keep your reasoning concise.
\end{quote}

\end{document}